\pdfoutput=1
\documentclass[11pt]{article}

\usepackage{acl}

\usepackage{times}
\usepackage{latexsym}

\usepackage{algorithm}

\usepackage{multicol}
\usepackage{multirow}
\usepackage{xcolor}
\usepackage{mathtools}
\usepackage{enumitem}
\usepackage[noend]{algpseudocode}
\usepackage{graphicx}
\usepackage{subcaption}
\usepackage{color}
\usepackage{colortbl}
\usepackage{amssymb}
\usepackage{booktabs}
\usepackage{multirow}
\usepackage{bbm}

\usepackage[T1]{fontenc}
\usepackage[utf8]{inputenc}

\usepackage{microtype}

\usepackage{inconsolata}

\title{Towards Robust and Generalized Parameter-Efficient Fine-Tuning \\ for Noisy Label Learning}

\author{Yeachan Kim$^{1}$\thanks{\ \ These authors contributed equally to this work.}, Junho Kim$^{1}$\footnotemark[1], SangKeun Lee$^{1,2}$ \\
$^{1}$Department of Artificial Intelligence, Korea University, Seoul, South Korea \\
$^{2}$Department of Computer Science and Engineering, Korea University, Seoul, South Korea \\
  \texttt{\{yeachan,monocrat,yalphy\}@korea.ac.kr}}

\begin{document}
\maketitle
\begin{abstract}

Parameter-efficient fine-tuning (PEFT) has enabled the efficient optimization of cumbersome language models in real-world settings. However, as datasets in such environments often contain noisy labels that adversely affect performance, PEFT methods are inevitably exposed to noisy labels. Despite this challenge, the adaptability of PEFT to noisy environments remains underexplored. To bridge this gap, we investigate various PEFT methods under noisy labels. 
Interestingly, our findings reveal that PEFT has difficulty in memorizing noisy labels due to its inherently limited capacity, resulting in robustness. However, we also find that such limited capacity simultaneously makes PEFT more vulnerable to interference of noisy labels, impeding the learning of clean samples. To address this issue, we propose \textbf{Clea}n \textbf{R}outing (CleaR), a novel routing-based PEFT approach that adaptively activates PEFT modules. In CleaR, PEFT modules are preferentially exposed to clean data while bypassing the noisy ones, thereby minimizing the noisy influence. To verify the efficacy of CleaR, we perform extensive experiments on diverse configurations of noisy labels. The results convincingly demonstrate that CleaR leads to substantially improved performance in noisy environments\footnote{Our code is available at \url{https://github.com/yeachan-kr/clear}}. 

% existing PEFT methods have overlooked the existence of noisy labels in real-world environments, adversely affecting generalization performance 

% datasets in such environments often contain noisy labels, adversely affecting generalization performance. However, such a close association within practical environments inevitably exposes PEFT methods to noisy labels, the effectiveness of PEFT in these conditions has not yet been explored.

% In this work, we investigate the PEFT method when exposed to noisy labels. Interestingly, our findings indicate that PEFT methods struggle to memorize noisy labels due to their inherently limited capacity, which in turn contributes to their robustness. However, we also find that such struggling with noisy labels prevents PEFT methods from learning even correctly labeled data as well, potentially resulting in underfitting.  Given that learning sufficiently from clean samples is also a crucial aspect alongside robustness, we propose Clean Routing (CleaR), a novel routing-based PEFT approach that preferentially exposes PEFT modules to clean samples through the adaptive routing, thereby minimizing the influence of noisy samples on PEFT. To verify the efficacy of the proposed method, we perform comprehensive experiments on diverse configurations of noisy labels. The results demonstrate that CleaR substantially improves the generalization ability of PEFT while maintaining or even reinforcing the robustness.
\end{abstract}

% \section{Introduction (yc)}

\section{Introduction}

The ever-growing size of pre-trained language models (PLMs) has presented significant challenges in adapting these models to desired tasks. In response to this practical limitation, parameter-efficient fine-tuning (PEFT) has emerged as a promising strategy for real-world environments. Instead of fine-tuning all weights, PEFT optimizes only a minimal set of parameters (e.g., biases \cite{bitfit}, adapters \cite{adapter}, prompts \cite{ptuning_v2}, or low-rank matrices \cite{lora}), thereby drastically cutting down the computation and storage costs. Such efficiency has led PEFT methods to become the preferred standard approaches for applying PLMs in real-world contexts, such as federated learning \cite{kim2023client,DBLP:conf/acl/LiaoMM23} and continual learning \cite{ermis2022memory,razdaibiedina2022progressive}.

% However, efficiency alone is not sufficient for real-world environments, as datasets in such settings often contain noisy labels \cite{real_noise_dataset1,real_noise_dataset2} that is known for adversely affecting the performance. Considering such a close association with practical environments, PEFT methods are inevitably exposed to noisy labels during training, which adversely affects the generalization ability of PLMs.

% Despite of this , there is a lack of prior research on whether the effectiveness of PEFT methods can be generalized to noisy environments.

While PEFT enables the efficient optimization of PLMs in real-world settings, datasets in such environments often contain noisy labels (i.e., incorrectly-labeled samples) \cite{real_noise_dataset1,real_noise_dataset2}, which adversely affects the generalization capabilities of PLMs \cite{stgn}. Given such distinct characteristics of the practical environments, PEFT methods are inevitably exposed to noisy labels during the optimization phase. Despite this significant challenge, there is a lack of prior research on the general adaptability of PEFT methods to noisy label learning (NLL) scenarios.
%noisy environments.

% whether the effectiveness of PEFT methods can be generalized to noisy environments. 

%Despite these noise labels hurting the generalization capability of neural models, there is a lack of prior research on whether the effectiveness of PEFT methods can be generalized to noisy environments.
%This close association with practical environments implies that PEFT methods are inevitably exposed to noisy labels during training. Despite this challenge, there is a lack of prior research on whether the effectiveness of PEFT methods can be generalized to noisy environments. 

In this work, we bridge this research gap by exploring PEFT under noisy environments. 
Our results reveal that PEFT struggles in memorizing noisy labels due to its inherently limited capacity, which interestingly provides robustness\footnote{Following \cite{DBLP:conf/aaai/WangWPZ21}, we define {robustness} as the preservation of the performance under noisy labels.} to noisy labels. 
%Interestingly, our results reveal that PEFT enjoys a certain level of robustness\footnote{Following \cite{DBLP:conf/aaai/WangWPZ21}, we define {robustness} as the preservation of the performance under noisy labels.} to noisy labels, due to the difficulty in memorization given its inherently limited capacity. 
%However, we also find that such limited capacity simultaneously interferes with PEFT in learning clean samples, potentially leading to sub-optimal performance. 
%However, we also find that such a struggle with noisy labels simultaneously interferes with PEFT in learning clean samples, potentially leading to sub-optimal performance. 
However, we also find that such limited capacity simultaneously makes PEFT more susceptible to interference of noisy labels, which impedes learning ability for clean samples, potentially leading to sub-optimal performance.
This characteristic markedly contrasts with the behaviors in full fine-tuning, presenting the necessity of PEFT that steers its limited learning capacity towards clean samples.
%Our results reveal that, due to the inherently limited capacity \cite{peft_analysis}, PEFT has difficulty in memorizing noisy labels, which interestingly resulting in the robustness\footnote{Following \cite{DBLP:conf/aaai/WangWPZ21}, we define {robustness} as the preservation of the performance under noisy labels.} to noisy labels. 
%However, we also find that such a struggle with noisy labels interferes with the PEFT in learning clean samples as well, potentially leading to sub-optimal performance. This characteristic markedly contrasts with the behaviors in full fine-tuning, presenting the necessity of the PEFT method that steers its limited learning capacity towards clean samples.

% notably, PEFT reveals robustness against noisy labels but tends to underfit even correctly-labeled data, potentially leading to sub-optimal performance. 
% This characteristic markedly contrasts with the behaviors in full fine-tuning (i.g., non-robustness and overfit). 
% The above observations pose the necessity for improved PEFT methods that retain the benefit of robustness while enhancing learning capacity, thereby pushing the boundaries between robustness and generalization.
%
%highlights the need for improved PEFT methods that 
%This highlights the need for improved PEFT methods that retain the benefit of robustness while enhancing learning capacity, thereby pushing the boundaries between robustness and generalization.

\begin{figure*}[t]
\centering
\includegraphics[width=1\textwidth]{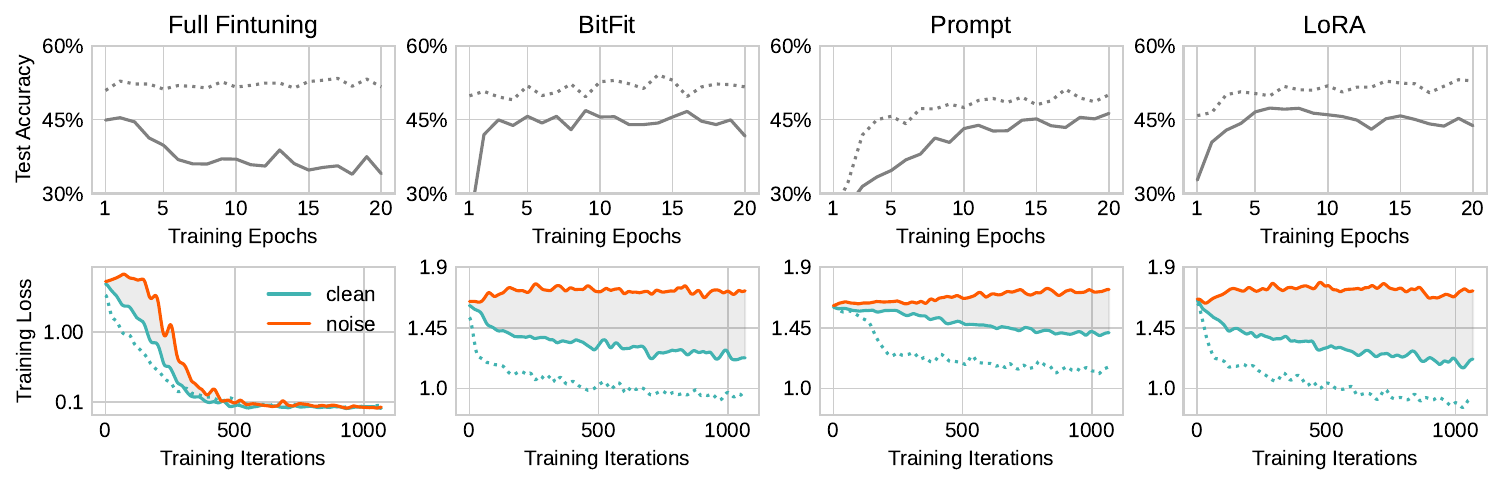}

\caption{Comparison between PEFT methods and full fine-tuning on SST-5 with symmetric noise (60\%). Dashed lines represent the training accuracy and loss of clean samples on uncorrupted datasets (i.e. only clean samples).}% for each fine-tuning method.}
%Plot (a) represents the test accuracy during the training. Plots (b) and (c) show positive and negative memorization.}% which indicates the memorization effect on clean samples and negative samples, respectively. }
\label{fig:peft_analysis}
\end{figure*}

In response, we propose \textbf{Clea}n \textbf{R}outing (CleaR), a novel routing-based PEFT approach that adaptively activates PEFT modules. Our main strategy is to preferentially expose PEFT modules to correctly-labeled samples, while bypassing PEFT weights for noisy samples, thereby minimizing the detrimental impact of noisy ones. To this end, CleaR estimates the probability that a given sample is correctly labeled through the lens of training dynamics. These probabilities are then conditioned to draw the routing decision such that potentially clean samples are more encouraged to route through the PEFT modules. The independent sampling across layers enables the fine-grained optimization for the PEFT modules. Consequently, engaging PEFT primarily with clean samples minimizes the influence of noisy samples on PEFT.
%\footnote{In this paper, we use the terms \textit{clean} and \textit{correctly-labeled} interchangeably to refer samples with ground truth annotation.}

The CleaR approach is designed to be model-agnostic, allowing us to integrate the concept of CleaR with various types of PEFT methods. To evaluate the effectiveness of CleaR-based PEFT methods, we conduct comprehensive experiments across diverse configurations, such as noise rate and noise type. Furthermore, we explore whether existing robust methods can be enhanced by seamlessly incorporating the CleaR approach. In summary, the contributions of this paper include the following:
\begin{itemize}%[leftmargin=*]
    \item We first explore the effectiveness of PEFT methods in the context of noisy environments, providing a comprehensive analysis of their robustness and limitations.
    
    \item We propose CleaR, a novel PEFT approach that adaptively activates the PEFT modules to improve generalization capability while minimizing the influence of noisy samples.
    
    \item We demonstrate that CleaR-based PEFT methods achieve superior performance across various NLP tasks even under heavy noise conditions, thereby pushing the boundaries of robustness and generalization.
\end{itemize}

\begin{figure*}[ht]
\centering
\includegraphics[width=1\textwidth]{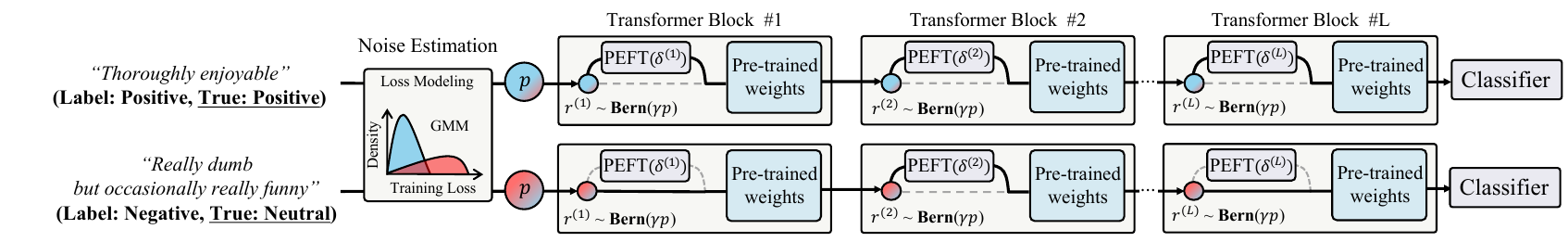}
% \hfill
\caption{Overview of the \textbf{Clea}n \textbf{R}outing. CleaR first estimate the probability of each sample being clean based on the training losses. Based on the estimated probability, CleaR adaptively activates PEFT modules by favoring the potentially clean samples.}

\label{fig:main_model}
\end{figure*}

\section{Investigation of PEFT on Noisy Labels}

In this section, we systematically investigate PEFT methods in the presence of noisy labels.

\paragraph{Noisy environment}Following the previous work \cite{stgn}, we simulate the noisy environment by randomly flipping the given labels. Specifically, we employ the SST-5 dataset with a symmetric noise rate of 60\% (i.e., 60\% of the training set contains incorrect labels). Note that the test set is not corrupted to confirm the generalization ability of the trained model. The detailed process is described in Appendix \ref{appendix:noise_process}.

\paragraph{PEFT methods} We analyze the three representative types of PEFT methods \cite{DBLP:conf/emnlp/Chen0ML22} with the full fine-tuning: \textbf{LoRA} \cite{lora} that adds trainable decomposition matrices; \textbf{BitFit} \cite{bitfit} that trains only biases; \textbf{Prompt Tuning} \cite{ptuning_v2} that appends learnable embeddings to the input of each layer\footnote{The setups for each PEFT methods are described in \S\ref{sec:C}}.

\paragraph{Observations} Figure \ref{fig:peft_analysis} presents the evaluation results of the PEFT methods alongside the full fine-tuning. The accuracy results (first row in the figure) show that PEFT methods reveal superior robustness to the full fine-tuning, even though all methods suffer from performance degradation. 
To gain further insights into the behavior of each method, we include the training loss for both clean\footnote{In this paper, we use the terms \textit{clean} and \textit{correctly-labeled} interchangeably to refer samples with ground truth annotation.} and noisy samples (second low in the figure), which enables to analyze the learning capacity on clean and noisy samples \cite{clean_first}. 
These results show that PEFT methods have difficulty in memorizing the noisy samples, which interestingly contributes to the robustness. However, it is evident that PEFT methods also face challenges in learning from clean samples (i.e., a large gap in losses of clean samples when exposed to the noisy dataset). 
%This result implies that PEFT inherently has a limited capacity \cite{peft_analysis}, and the struggle with noisy labels impedes its ability in learning from clean data. 
The result implies that PEFT, which inherently has limited capacity \cite{peft_analysis}, is more vulnerable to the interference of noisy labels, impairing its learning ability on clean data.
This limitation can lead to sub-optimal performance, underscoring the need for PEFT methods that can steer its limited capacity toward clean samples.

\section{{CleaR}: PEFT with Clean Routing}

In this section, we elaborate \textbf{Clea}n \textbf{R}outing (CleaR) in detail. The core idea is to adaptively activate the PEFT modules to circumvent the detrimental effect from noisy labels. To achieve this, CleaR estimates the probability whether a given sample is correctly labeled, leveraging the distinct training dynamics between clean and noisy samples. With these probabilities, CleaR steers the potentially clean samples to route through PEFT modules, whereas the noisy ones are directed to bypass PEFT modules, thereby minimizing their influence on the PEFT. To improve the routing stability of CleaR, we also introduce consistency regularization for PEFT modules. Figure \ref{fig:main_model} illustrates the overall procedures of CleaR.

\subsection{Parameter-Efficient Fine-Tuning Modules}\label{subsect:peft}

We start by defining PEFT modules in CleaR. Since CleaR is designed to be module-agnostic, diverse PEFT methods can be seamlessly integrated with CleaR method. To showcase such applicability, we consider four representative PEFT modules (i.e., Adapter, BitFit, Prompt-tuning, and LoRA), which are commonly employed within NLP community\footnote{The detailed illustrations of these modules on CleaR are shown in the Appendix \S \ref{sec:C} (Figure \ref{fig:clear_model_detail})}. While these PEFT modules have distinct characteristics, they can be succinctly represented as additional parameters $\boldsymbol{\delta}$ added to the pre-trained weights $\boldsymbol{\theta}$ of PLMs \cite{peft_analysis}. 
More specifically, since PEFT modules are uniformly distributed across all layers, we represent these modules as a set of additional parameters $\boldsymbol{\delta} = \{\delta^{(1)}, \delta^{(2)}, \dots, \delta^{(L)}\}$, where $L$ is the number of layers. 
Let the prediction involving these PEFT modules be denoted as $f(x, \boldsymbol{\delta} +\boldsymbol{\theta})$, the objective with an arbitrary loss function $\mathcal{L}$ can be formulated as follows:
\begin{equation}
    \underset{{\delta}}{\text{min}} \; \mathcal{L}(x) = \mathcal{L}(f(x, \boldsymbol{\delta} + \boldsymbol{\theta}), y),
\end{equation}
where $x$ and $y$ denote the training sample and the given label, respectively. Note that the PEFT modules and the task-specific classifier are only updated during training. 

% %We train our model for k epochs after warm-up and estimate the 
\subsection{Routing PEFT Modules}\label{subsect:clear}
Building upon these PEFT modules, we introduce a clean routing scheme that adaptively activates modules according to the estimated probability that a given sample is correctly labeled.

\paragraph{Estimating clean probability for routing} 
To derive the clean probability for each sample, we leverage the distinct learning patterns when learning with clean and noisy samples: \textit{deep networks prefer to learn clean samples first before fitting noisy ones} \cite{clean_first}. Namely, noisy samples tend to have a higher loss than clean samples in the early training stage. 
This enables to distinguish potentially clean samples from the datasets based on loss deviation \cite{mentornet,co-teaching}. 
Taking advantage of such phenomena, we adopt the widely-used Gaussian Mixture Model (GMM) in noise label learning \cite{devidemix, selfmix}, in which the probability of samples being clean is estimated by the per-sample loss. 

%distinguish clean samples by feeding the per-sample loss. %using the Expectation Maximization algorithm. 
Based on the estimated mixture models, we compute the clean probability $p$ using the posterior probability, i.e., $p(g|\ell)$ where $\ell$ is the loss for the training sample, and $g$ is the Gaussian component with a smaller mean (i.e., smaller loss). Specifically, we first train our model for the $k$ epochs warm-up to measure the loss of samples, and then we estimate the clean probability for each training sample on every subsequent epoch. 
%It is noteworthy that we leverage the training losses obtained from the previous epoch for estimating clean probability to reduce the additional computational costs from redundant forward passes.
It is noteworthy that we leverage training losses obtained from the previous epoch to estimate clean probabilities, thereby reducing the additional computational cost from redundant forward passes.
%It is noteworthy that we use the training losses obtained from the previous epoch, as calculating the training loss on-the-fly before the routing introduces a redundant forward pass.

\paragraph{Sampling routing decision}
Once the clean probability is estimated, the PEFT modules are stochastically routed across the transformer layers. To derive the routing decision (i.e., routing through PEFT or bypassing PEFT), we sample the decision from a Bernoulli distribution with the estimated clean probability $p$:

\begin{equation}\label{eq:sampling}
r \sim \textsc{Bernoulli}(\gamma p),
\end{equation}
where $r$ is an independent Bernoulli random variable with a probability $\gamma p$ of being $1$ and a probability $1 - \gamma p$ of being 0. The coefficient $\gamma \in [0,1]$ limits the range of clean probability, setting its upper bound at $\gamma$. This coefficient plays a role in preventing over-reliance on the estimated probability, considering that small-loss samples might still contain noisy samples (i.e., high clean probability despite being noisy samples) \cite{devidemix}. 
Crucially, this routing decision is independently made at each layer, allowing for the fine-grained differentiation of each sample's influence based on its probability of being clean.
For example, if the clean probability is 70\% and the number of layers is 10, seven PEFT modules are activated in average\footnote{While different positions of the PEFT could have varying effects on the prediction \cite{design}, in this work, we focus solely on the clean probability as a trigger to activate the PEFT. We reserve further exploration of this for future work.}.

\paragraph{Activating PEFT based on the decision}
The routing decisions across different layers are then applied to all PEFT modules. Formally, let the hidden states in the $l$-th layer be denoted as $h^{(l)}$, and the hidden state in the next layer is derived as follows:
\begin{equation}
h^{(l+1)} = \begin{cases}
\text{Trans}^{(l)} (h^{(l)}, \delta^{(l)} + \theta^{(l)}),&\text{if } r^{(l)}=1 \\ \text{Trans}^{(l)} (h^{(l)}, \theta^{(l)}),&\text{if }r^{(l)}=0
\end{cases} 
\end{equation}
where $\delta^{(l)}$ and $\theta^{(l)}$ represent PEFT module and pre-trained parameters in the $l$-th layer, respectively, and $r^{(l)}$ indicates the routing decision on the layer. $\text{Trans}^{(l)}(\cdot)$ denotes the function of the transformer block. Through the above routing decision, CleaR activates a subset of PEFT modules, i.e., $\boldsymbol{{\delta}_r} = \{\delta^{(l)} | \delta^{(l)} \in \boldsymbol{\delta}, r^{(l)} = 1\}$, on each forward pass. %이러한 Routing을 통해, 최종적으로 \delta_r = {}을 sample 하여 모델을 학습한다.
%, such as multi-head self-attention and position-wise feed-forward networks.
This routing scheme ensures that PEFT modules are favorably activated for potentially clean samples and deactivated for noisy ones, thereby reducing the influence of noisy samples. 

\paragraph{CleaR in inference phase} 
As clean routing only performs in training, we need to decide the routing strategy during inference. To make the most of well-trained PEFT modules and ensure consistency with training, we empirically set the routing probability to the upper bound, i.e., $p = 1.0$ in Eq.~\eqref{eq:sampling}, and observe that it works well in practice.

\subsection{Consistency Regularization for CleaR}\label{subsect:cr}

While the routing scheme effectively mitigates the influence of noisy labels, model predictions may end up being overly diverse due to varying activations with each forward pass, potentially resulting in training instability. % \cite{modular_survey}. 
To address this issue, we introduce a consistency regularization to minimize the model variability. Considering that guiding the model to adhere to past predictions can enhance the stability and consistency of training \cite{consistency_cvpr, consistency_bert}, we regulate the model by minimizing the distance between its current and previous predictions. Specifically, we make ensemble predictions from multiple forwards to reduce predictive variance and increase stability:
\begin{equation}
    f_{\text{ens}}(x, \bar{\boldsymbol{\delta}}_{r} + \theta) = \frac{1}{N} \sum_{k=1}^{N} f(x, \bar{\boldsymbol{\delta}}_{r,k} + \boldsymbol{\theta}),
\end{equation}
%where $\bar{\boldsymbol{\delta}}_r$ represents PEFT modules trained in the previous epoch, and $N$ is the number of forward passes. 
where $N$ is the number of forwards, and $\bar{\boldsymbol{\delta}}_{r,k}$ represents activated PEFT modules in the $k$-th forward of the previously trained model. It is noteworthy that, for computational efficiency, we reuse the predictions, which were previously used for fitting GMM. With the derived predictions, the model with CleaR is optimized with the following loss:
\begin{multline}
        \underset{{\delta_r}}{\text{min}} \; \mathcal{L}(x) = \mathcal{L}_{\text{CE}}(    f(x, \boldsymbol{\delta}_r + \boldsymbol{\theta}), y)  \\ + \lambda \mathcal{L}_{\text{CE}}(f(x, \boldsymbol{\delta}_r + \boldsymbol{\theta}), f_{\text{ens}}(x, \bar{\boldsymbol{\delta}}_r + \boldsymbol{\theta})).
\end{multline}
where $\mathcal{L}_{\text{CE}}(\cdot)$ indicates the cross-entropy loss, and $\lambda$ is a coefficient to control the strength of the regularization.

\begin{table*}[t]
\centering
\caption{Evaluation results of Peak accuracy and Average accuracy on SST-5 test set under different levels of label noise. The best and second best results are highlighted in \textbf{boldface} and \underline{underlined}, respectively.}
\resizebox{\textwidth}{!}{
\begin{tabular}{@{}lcclclclclclcl@{}}
\toprule
\multirow{3}{*}{Methods} & \multirow{3}{*}{Clean} & \multicolumn{6}{c}{Symmetric}                                                        & \multicolumn{6}{c}{Asymmetric}                                                       \\ \cmidrule(l){3-8} \cmidrule(l){9-14}
                         &                        & \multicolumn{2}{c}{20\%} & \multicolumn{2}{c}{40\%} & \multicolumn{2}{c}{60\%} & \multicolumn{2}{c}{10\%} & \multicolumn{2}{c}{20\%} & \multicolumn{2}{c}{40\%} \\ 
                         &                        & Peak. & Avg. & Peak. & Avg. & Peak. & Avg. & Peak. & Avg. & Peak. & Avg. & Peak. & Avg. \\ \midrule
Full Fine-tuning         & \underline{53.4}        & 51.3            & 47.0       & 50.6        & 42.9        & 47.9        & 35.5    & \textbf{52.5}         & 49.1    & 50.8        & 46.5       & 46.1         & 37.4              \\ \midrule

\multicolumn{7}{l}{\textit{PEFT methods}}             &     &     &     &   &       &          \\ 

Adapter \citeyearpar{adapter}                 & 53.3                    & 51.9            & 48.1       & 50.5        & 45.8        & 47.2         & 38.1    & \underline{52.2}         & 51.0     & 50.9        & 47.0       & 48.1        & 38.0           \\ 
BitFit \citeyearpar{bitfit}                  & 53.0                    & 51.7            & 51.0       & 50.8        & 48.1        & 48.1        & 43.5    & 52.1         & 50.5     & \underline{52.1}        & 49.2       & \underline{48.9}         & 42.1               \\
Prompt \citeyearpar{ptuning_v2}                  & 52.7                    & 51.1            & 48.6       & 50.7        & 49.1        & 47.7        & 45.7     & 51.7         & 50.8    & 49.4        & 48.2       & 46.1         & 41.7              \\
LoRA \citeyearpar{lora}                    & \textbf{53.6}           & \underline{52.0}& 49.5       & 50.2        & 47.5        & 48.2        & 46.1     & 51.9         & 51.1    & 50.5        & 47.4       & 47.2         & 41.8             \\ \midrule

\multicolumn{7}{l}{\textit{PEFT methods with CleaR (ours)}}             &     &     &     &   &       &          \\ 

CleaR$_{\text{Adapter}}$            & \underline{53.4}        & \textbf{52.4}   & \textbf{51.8}  & \underline{51.5}  & \underline{50.4}   & \underline{50.4}        & \underline{49.7}    & \textbf{52.5}         & 50.8     & 51.4         & 47.4        & 48.1         & 44.6              \\
CleaR$_{\text{BitFit}}$             & 53.1                    & 51.9            & \underline{51.1}   & \textbf{51.6}  & \textbf{51.2}   & \textbf{51.4}       & \textbf{51.1}    & 52.0         & \textbf{51.4}     & \textbf{52.3}         & \underline{50.4}        & \textbf{49.2}         & \textbf{48.3}         \\
CleaR$_{\text{Prompt}}$             & 52.6                    & 51.0            & 50.5       & 51.4        & 49.5        & 49.4         & 47.2        & 52.1         & \underline{51.2}     & 52.0         & \textbf{51.4}      & 47.8         & \underline{46.5}                       \\
CleaR$_{\text{LoRA}}$               & 53.3                    & 51.4            & 50.1       & 51.2        & 49.0        & 50.0        & 48.9  & 52.0         & 51.1     & 51.0         & \underline{50.4}        & 47.6         & 43.2              \\ \bottomrule
\end{tabular}
}
\label{tab:sst}
\end{table*}

%Experiment 전체 RQ로 시작(...)? or Analysis부터 RQ로 시작(SelfMix 논문 전개방식)? 
\section{Experiments}
%We evaluate CleaR across diverse configurations of noisy labels. We specifically demonstrate that CleaR significantly improves the generalization capability of PEFT while preserving its robustness to noisy labels. 
We demonstrate the efficacy of CleaR across diverse configurations of noisy environments.

\subsection{Configurations of Noisy Labels}
To comprehensively assess our method diverse scenarios characterized by different noisy labels, we evaluate each baseline against three distinct types of noisy labels: symmetric, asymmetric, and instance-dependent. We provide detailed descriptions of each noisy label type and the methodology for constructing noisy label datasets in \S \ref{appendix:noise_process}.

\subsection{Baselines and Implementations} 
Following the previous works, we use the BERT-base and BERT-large model \cite{bert}. Building on this PLM, we mainly compare CleaR with the full fine-tuning and widely-used PEFT methods including Adapter \cite{adapter}, LoRA \cite{lora}, Prompt tuning \cite{ptuning_v2}, and BitFit \cite{bitfit}. For a fair comparison, we utilize the same settings for PEFT modules (e.g., bottleneck dimension, prompt length) for baselines and our CleaR. The detailed implementations are represented in \S \ref{appendix:implement}. Additionally, we compare CleaR with the existing NLL methods to confirm the competitiveness of the proposed method in \S\ref{sec:5.3}. 

\subsection{Evaluation Metric}
Building upon previous work \cite{stgn}, we assess each baseline model using two metrics: the instantaneous peak accuracy and the average accuracy across the last few epochs. The former metric evaluates the model's generalization performance, while the latter reflects its stability. Consequently, a smaller gap between these two metrics indicates a more effective model to noisy labels. For all CleaR models, we report the average performance on 10 different seeds considering their stochasticity.
%Therefore, as the gap between the two metrics is narrow, the model can be considered to be robust.

\begin{table*}[t]
\centering
\caption{Evaluation results of Peak accuracy and Average accuracy on BANKING77 test set under different levels of label noise. The best and second best results are highlighted in \textbf{boldface} and \underline{underlined}, respectively.}
\resizebox{\textwidth}{!}{
\begin{tabular}{@{}lcclclclclclcl@{}}
\toprule
\multirow{3}{*}{Methods} & \multirow{3}{*}{Clean} & \multicolumn{6}{c}{Symmetric}                                                        & \multicolumn{6}{c}{Asymmetric}                                                       \\ \cmidrule(l){3-8} \cmidrule(l){9-14}
                         &                        & \multicolumn{2}{c}{20\%} & \multicolumn{2}{c}{40\%} & \multicolumn{2}{c}{60\%} & \multicolumn{2}{c}{10\%} & \multicolumn{2}{c}{20\%} & \multicolumn{2}{c}{40\%} \\ 
                         &                        & Peak. & Avg. & Peak. & Avg. & Peak. & Avg. & Peak. & Avg. & Peak. & Avg. & Peak. & Avg. \\ \midrule
Full Fine-tuning         & 92.9                   & 88.8        & 83.4       & 84.3        & 72.6       & 78.2        & 58.5      & 90.8        & 87.6    & 87.3        & 79.4       & 66.9        & 54.6            \\ \midrule
\multicolumn{7}{l}{\textit{PEFT methods}}             &     &     &     &   &       &          \\ 

Adapter \citeyearpar{adapter}                 & 92.7                   & 88.5        & 85.4       & 86.6        & 78.4       & 80.9        & 67.1      & 90.3        & 88.6    & 86.7        & 78.5       & 65.3        & 56.2            \\ 
BitFit \citeyearpar{bitfit}                  & 92.5                   & 88.9        & 88.7       & 86.7        & 85.9       & 80.1        & 76.5      & 90.2        & 89.8    & 86.3         & 83.1       & 66.7        & 62.4            \\
Prompt \citeyearpar{ptuning_v2}                   & 91.9                   & 87.8        & 87.4       & 85.6         & 84.5       & 83.2        & 77.2     & 89.7        & 88.4     & 85.4        & 84.9       & 61.6         & 58.9          \\
LoRA \citeyearpar{lora}                     & \underline{93.0}     & 89.2        & 88.3       & 86.8        & 85.8       & 81.9        & 77.5      & 90.1        & 88.6    & 86.9        & 83.1       & 64.5        & 61.8            \\ \midrule

\multicolumn{7}{l}{\textit{PEFT methods with CleaR (ours)}}             &     &     &     &   &       &          \\ 

CleaR$_{\text{Adapter}}$            & \textbf{93.1}       & \textbf{90.1}         & \underline{89.7}        & \textbf{88.2}         & \textbf{87.3}        & 82.3    & 80.2      & \textbf{91.4}        & 90.3     & \textbf{87.6}         & \textbf{86.1}        & \underline{67.3}         & \underline{66.1}              \\
CleaR$_{\text{BitFit}}$             & 92.4                    & 89.8         & 89.2       & 87.3         & \underline{86.9}        & 82.9    & \underline{82.2}       & 90.7        & \textbf{90.4}    & \underline{87.5}         & 86.1        & 67.1         & 63.4          \\
CleaR$_{\text{Prompt}}$             & 92.1                    & 88.1         & 87.6        & 85.8         & 84.9        & \underline{83.7}   & 81.0       & 89.9        & 89.2     & 85.7         & 84.8        & 64.5         & 62.3               \\
CleaR$_{\text{LoRA}}$               & 92.8                    & \underline{90.0}         & \textbf{89.8}        & \textbf{87.4}         & \underline{86.9}        & \textbf{84.2}     & \textbf{83.5}     & \underline{91.3}        & \underline{90.3}   & 87.2         & \underline{85.9}        & \textbf{68.9}         & \textbf{68.1}          \\ \bottomrule
\end{tabular}
}
\label{tab:bank}
\end{table*}

\subsection{Sentiment Analysis}
We first evaluate baselines in sentiment analysis due to the inherent subjectivity of this task, which often results in noisy labels.
Following the previous work, we use the Standard Sentiment Treebank (SST-5) dataset \cite{sst}. %including fine-grained sentiment classes. 
For the levels of noisy labels, we scale the symmetric noise from 20\% to 60\% and asymmetric noise from 10\% to 40\%, respectively.

Table \ref{tab:sst} presents the evaluation results for the task. As observed in the previous analysis, PEFT exhibits better robustness than full fine-tuning across different types and levels of noisy labels. Notably, CleaR-based methods show substantial improvement compared to PEFT methods on both metrics. This demonstrates that CleaR enhances the generalization capability of PEFT (i.e., improved peak accuracy) while maintaining or even strengthening its robustness (i.e., reduced gaps between peak and average accuracy). 

\begin{table}[t]
\centering
\caption{Ablation study of CleaR on SST-5 (60\% of symmetric noise). For the ablation of routing strategies, we remove the consistency regularization to solely evaluate each routing strategy.}
\resizebox{\columnwidth}{!}{
\begin{tabular}{@{}lcc@{}}
\toprule
Methods                              & Peak. & Avg. \\ \midrule
CleaR$_{\text{Adapter}}$(ours)                         & 50.4 & 49.7    \\ \midrule
\multicolumn{3}{c}{\cellcolor[gray]{.9}Components in CleaR} \\
CleaR w/o Clean Routing                    & 48.4 & 41.1    \\
CleaR w/o Regularization          & 49.9 & 48.6    \\
CleaR w/o Clean Routing \& Regularization & 47.2 & 40.0    \\ \midrule
\multicolumn{3}{c}{\cellcolor[gray]{.9}Routing Strategy in CleaR} \\
CleaR w/ Clean Routing & 49.9 & 48.6    \\
CleaR w/ Deterministic Routing & 48.1 & 44.2    \\
CleaR w/ Random Routing & 47.5 & 40.5    \\
CleaR w/ Noisy Routing & 46.9 & 34.3    \\ \bottomrule
\end{tabular}
}
\label{tab:ablation}
\end{table}

\subsection{Intent Detection}
We further evaluate CleaR on intent detection. Given that the task is typically employed in conversational systems, the query usually consists of only a few words (e.g., 6 to 12 words) \cite{banking77}. Such brevity can amplify ambiguity, potentially leading to noisy annotations. We utilize the BANKING77 \cite{banking77} that encompasses 77 fine-grained intent categories. For the levels of noisy labels, we scale the symmetric noise from 20\% to 60\% and asymmetric noise from 10\% to 40\%, respectively.

Table \ref{tab:bank} presents the evaluation results for intent detection. Similar to the previous task, PEFT achieves superior robustness to noisy labels compared to full fine-tuning by showing higher average accuracy even on the highest noisy levels. Meanwhile, CleaR consistently outperforms both the PEFT and full fine-tuning methods, demonstrating its efficacy to improve the robustness. It is noteworthy that in certain setups, while PEFT shows better average accuracy, it yields slightly lower peak accuracy due to its limited capacity to memorize clean samples (e.g., asymmetric noise levels between 20\% and 40\%). Interestingly, CleaR significantly boosts the peak accuracy of each PEFT variant, achieving similar or even better peak accuracy than full fine-tuning. These results again verify that CleaR successfully mitigates the limited memorization of PEFT methods, thereby leading to better generalization and robustness. 

\section{Analysis}
To make a more comprehensive analysis of our CleaR, we designed a series of fine-grained experiments aimed at addressing the following research questions (RQs):
\setlist{nolistsep}
\begin{itemize}[leftmargin=*, noitemsep]
\item \textbf{RQ1.} \,\,How does each component within CleaR contribute to its overall performance? (\S \ref{sec:5.1})
\item \textbf{RQ2.} Can CleaR learn clean samples well while minimizing the influence of noisy ones? (\S \ref{sec:5.2})
%\item \textbf{RQ2.} \,\,Can CleaR be integrated with other methods for learning with noisy labels? (\S \ref{sec:5.2})
\item \textbf{RQ3.} \,\,Can CleaR be combined with other noisy label learning methods? (\S \ref{sec:5.3})
\item \textbf{RQ4.} Does CleaR offer improvements under more realistic noisy label scenarios? (\S \ref{sec:5.4})
\item \textbf{RQ5.} Can CleaR be generalized to the large-sized model? (\S \ref{sec:B})
\end{itemize}

\begin{figure}[t]
\centering
\includegraphics[width=\linewidth]{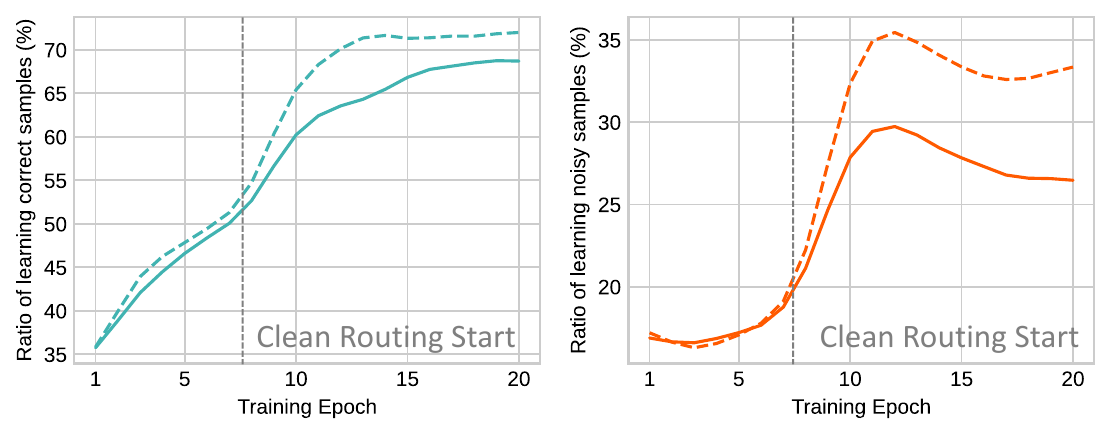}
% \hfill
\caption{Ratios of memorizing clean (\textbf{Left}, larger is better) and noisy samples (\textbf{Right}, smaller is better) on different routing methods. Dashed lines and solid lines indicate Deterministic Routing and Clean Routing (ours), respectively.}
%indicate Deterministic Routing and solid lines indicate Clean Routing (ours).}
\label{fig:mem_diff}
\end{figure}

\begin{figure}[t]
\centering
\subfloat[Impact on learning clean samples]{%
  \includegraphics[width=\linewidth]{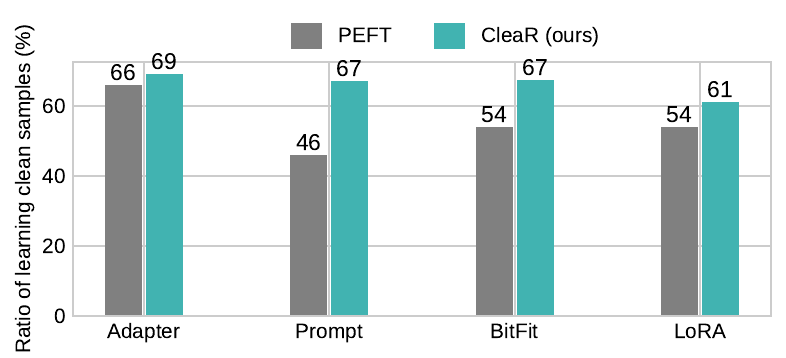}%
  % \label{exp:mot:info} 
}

\subfloat[Impact on learning noisy samples]{%
  \includegraphics[width=\linewidth]{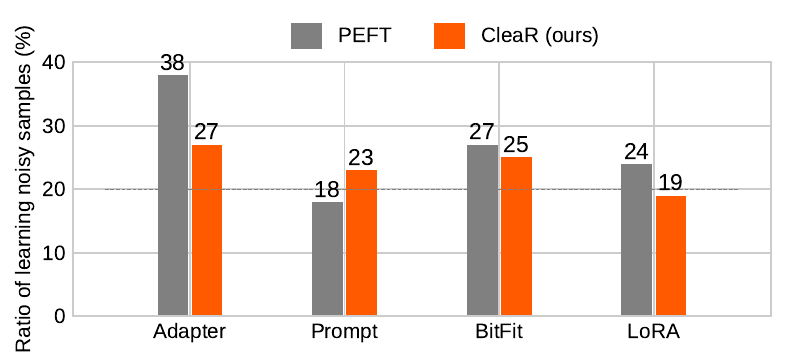}%
  % \label{exp:mot:acc}
}
% \hfill
\caption{Impact on two memorizations when applying CleaR to PEFT methods. Best viewed in color..}
\label{fig:clear_before_after}
% \vspace*{-0.09in}
\end{figure}

\subsection{Ablation Studies on CleaR (RQ1)}
\label{sec:5.1}
We perform ablation studies to investigate the contributions of each component and routing strategy in CleaR. Table \ref{tab:ablation} presents the ablation results. %Here, we perform the analysis on CleaR$_{\text{Adapter}}$ due to its general applicability.

\paragraph{Routing and Regularization} As shown in the upper part of Table 3, omitting routing and consistency regularization largely affects both peak and average accuracy. The results demonstrate that each component is essential for improving generalization and robustness. In particular, we observe that the routing mechanism plays a significant role.

\begin{table}[t]
\centering
\caption{Complementary effect of incorporating CleaR to NLL methods on SST-5 (60\% symmetric noise).}
\resizebox{\columnwidth}{!}{
\begin{tabular}{@{}cccc@{}}
\toprule
% \multirow{2}{*}{PEFT Methods}                                                & \multirow{2}{*}{Methods} & \multicolumn{2}{c}{SST-5 ($r = 60\%$)}       \\
   Methods                                                                          & NLL Methods                          & \multicolumn{1}{l}{Peak.} & \multicolumn{1}{l}{Avg.} \\ \midrule
\multirow{4}{*}{Full Fine-tuning}                                                        & None              & 47.9                    & 35.5                        \\
                                                                             & Co-teaching \citeyearpar{co-teaching}             & 50.1                     & 46.1                        \\
                                                                             & SELC \citeyearpar{selc}                     & 48.5                     & 39.7                        \\
                                                                             & STGN \citeyearpar{stgn}                    & 47.7                     & 38.6                        \\ \midrule
\multirow{4}{*}{Adapter}                                                        & None              & 47.2                     & 38.1                        \\
                                                                             & Co-teaching \citeyearpar{co-teaching}             & 50.3                     & 45.9                        \\
                                                                             & SELC \citeyearpar{selc}                    & 47.5                     & 39.7                        \\
                                                                             & STGN \citeyearpar{stgn}                    & 48.7                     & 39.8                        \\ \midrule
\multirow{4}{*}{\begin{tabular}[c]{@{}c@{}}CleaR$_{\text{Adapter}}$ \\ (ours)\end{tabular}} & None              & 50.4                     & 49.7                        \\
                                                                             & Co-teaching \citeyearpar{co-teaching}             & \underline{50.6}                     & \underline{50.1}                        \\
                                                                             & SELC \citeyearpar{selc}                    & 50.5                     & \textbf{50.2}                        \\
                                                                             & STGN \citeyearpar{stgn}                     &   \textbf{50.8}                   & 49.4                        \\ \bottomrule
\end{tabular}
}
\label{tab:nll}
\end{table}

\paragraph{Routing strategy} To further investigate the proposed routing scheme, we compared it with three different schemes: (i) Random Routing, where PEFT modules are activated randomly; (ii) Noisy Routing, where PEFT modules are activated in proportion to the noisy probability, representing the opposite approach of CleaR; and (iii) Deterministic Routing, where the noisy samples are filtered out by leveraging the estimated probability based on GMM, and PEFT modules are only deterministically activated on remaining samples.
%filter out the noisy samples by leveraging the estimated probability based on the GMM, and only deterministically activate the PEFT modules on remain samples. %PEFT modules are only deterministically activated on remain samples.

As shown in the lower part of Table 3, adopting Random and Noisy Routing substantially decreases both metrics, indicating that favoring clean samples in the routing is indeed beneficial to achieve both generalization and robustness ability.
We then compare our Clean Routing with the Deterministic Routing. We observe that Deterministic Routing exhibits inferior performance than our Clean Routing. Moreover, Figure \ref{fig:mem_diff} illustrates the differences in accuracy on clean and noisy samples between Clean Routing and Deterministic Routing. 
It reveals that Deterministic Routing leads to increased memorization of noisy samples compared to our stochastic Clean Routing, which is attributed to lower performance. These results emphasize the significance of stochastic routing in Clean Routing, which can differentiate samples in a more fine-grained manner.

%\subsection{Positive and Negative Memorization}
\subsection{Memorization Effects (RQ2)}
\label{sec:5.2}

To confirm whether CleaR mitigates the underfitting problems of PEFT methods on clean samples, we compare the ratio of memorizing clean and noisy samples after fine-tuning.
%positive and negative memorization after fine-tuning. 
Figure \ref{fig:clear_before_after} shows the comparison. Notably, we observe that adopting CleaR into PEFT methods largely increases the memorization of clean samples, while preserving or even reducing the memorization of noisy samples. These results verify that CleaR successfully mitigates the underfitting of PEFT methods on the clean dataset by favorably activating the PEFT modules on potentially clean samples.

\subsection{CleaR on Different NLL Methods (RQ3)}
\label{sec:5.3}
%To investigate whether our framework can further enhance the other NLL approaches, 
We compare CleaR with existing NLL methods to demonstrate its applicability. We consider three approaches: Co-teaching \cite{co-teaching}, SELC \cite{selc}, and STGN \cite{stgn}\footnote{The detailed settings are included in the Appendix \S \ref{appendix:nll}}. Table \ref{tab:nll} showcases the comparison and integration results with CleaR and Adapter.
We observe that NLL methods can be enhanced with CleaR, as it can be seamlessly integrated by adding PEFT modules. 
The results show that while combining NLL methods with Adapter brings marginal improvement, adopting CleaR leads to a marked enhancement across both metrics. This suggests that CleaR can place the current state-of-the-art NLL methods on a more solid footing by enjoying the robustness and improved generalization ability of PEFT.

\subsection{Instance-dependent Label Noise (RQ4)}
\label{sec:5.4}
To investigate CleaR in more realistic settings, we evaluate CleaR in scenarios where noisy labels arise from input features. Table \ref{tab:instance} presents the evaluation results on SST-5 with the instance-dependent noise. Similar to the other noise settings, CleaR consistently yields the best performance on both peak and average accuracy across different noise ratios, highlighting the validity of CleaR in addressing feature-dependent noise.

\begin{table}[t]
\centering
\caption{Peak and Average accuracy (\%) on SST-5 under different levels of instance-dependent noise.}
\resizebox{\columnwidth}{!}{
\begin{tabular}{@{}lccccc@{}}
\toprule
\multirow{2}{*}{Method} & \multirow{2}{*}{Clean} & \multicolumn{2}{c}{40\%} & \multicolumn{2}{c}{60\%} \\ \cmidrule(l){3-6} 
 &  & Peak. & Avg. & Peak. & Avg. \\ \midrule
Full Fine-tuning & \underline{53.4} & 49.0 & 43.9 & 44.8 & 38.9 \\ \midrule
\multicolumn{6}{l}{\textit{PEFT methods}}                 \\ 
Adapter & 53.3 & 48.8 & 44.1 & 44.2 & 39.7 \\
BitFit & 53.0 & 50.0 & 45.4 & 44.8 & 41.6 \\
Prompt & 52.7 & 49.5 & 46.2 & 42.8 & 38.8 \\
LoRA & \textbf{53.6} & 49.1 & 46.9 & 44.2 & 39.8 \\ \midrule

\multicolumn{6}{l}{\textit{PEFT methods with CleaR (ours)}}                \\ 

CleaR$_{\text{Adapter}}$ & \underline{53.4} & \underline{50.5} & 46.4 & \underline{45.7} & \underline{43.4} \\
CleaR$_{\text{BitFit}}$ & 53.1 & \textbf{51.0} & \underline{46.5} & 45.2 & 42.6 \\
CleaR$_{\text{Prompt}}$ & 52.6 & 50.2 & 44.1 & 44.8 & 43.2 \\
CleaR$_{\text{LoRA}}$ & 53.3 & 49.7 & \textbf{47.1} & \textbf{46.5} & \textbf{44.8} \\ \bottomrule
\end{tabular}%
}
\label{tab:instance}
\end{table}

\section{Related Work}
\subsection{Robustness of PEFT Methods}
The well-known beneficial properties of PEFT are its generalization capabilities in low-data environments \cite{peft_few_shot, bitfit} and stability \cite{adapter, peft_stable}. These studies have demonstrated that full fine-tuning suffers from overfitting on small datasets, whereas PEFT methods achieve superior performance in few-shot settings due to the regularization effect on the pre-trained models \cite{peft_aaai_analysis}. 
Regarding model calibration, recent work has shown that preserving the pre-trained features through the PEFT methods improves the model calibration by preventing overfitting \cite{pretrain_calibration, peft_analysis}. Following the robustness on the overfitting, several studies have demonstrated that PEFT methods also being less suffered from catastrophic forgetting \cite{peft_forgetting}, and their robustness against forgetting is further evidenced in continual learning scenarios \cite{ermis2022memory}.

Previous studies have broadened the understanding of PEFT in diverse aspects. Building upon this research direction, we explore its robustness to noisy labels, which remains a challenging problem within the deep learning community. Therefore, exploring PEFT methods in this context could provide valuable insights into their practical applicability.

\subsection{Noisy Label Learning in NLP}

Noisy labels are inevitably introduced on large-scale datasets \cite{real_noise_dataset1,noisywiki}. 
To mitigate the influence of noisy labels in text classification, \citet{noise_model} have proposed a noise transition matrix on top of the classifier, learning the transition distribution of noisy labels, and \citet{noise_llm} have tackled noisy labels in classification tasks with supplemental guidance from large language models (e.g., ChatGPT). In addition, \citet{dygen} have utilized the dynamics of features with generative models during training to learn the transition matrix for noisy labels.
For named-entity recognition, \citet{noise_ner2} have introduced a self-training method based on the contextualized augmentations, and \citet{noise_ner} have proposed a co-regularization in which multiple models teach each other to avoid negative memorization. 
In entity linking, \citet{noise_entity_linking} have explored an interpretable method to identify clean samples on the premise that interpretable samples tend to be clean. 
Recently, \citet{stgn} have proposed a noise-robust optimization by introducing stochastic tailor-made gradient noise. 

While these works have achieved robustness to noisy labels, these works typically consider the scenarios that the entire parameters in PLMs are optimized during fine-tuning, which is challenging due to the huge number of parameters. To the best of our knowledge, we are the first to explore the PEFT methods for NLL. Given that CleaR is built upon PEFT methods, it can enjoy the benefits of training efficiency and robustness, which is favorable in the real-world environment.

\section{Conclusion}
In this work, we have explored whether the PEFT methods can be generalized to noisy environments. Interestingly, we have found that, while the limited capacity of PEFT allows the robustness to noisy labels, it can also act as a double-edged sword that hinders learning even with clean samples. In response, we have proposed CleaR, a novel routing-based PEFT approach that adaptively activates the PEFT modules by considering the probability of samples being clean. Extensive experiments have convincingly demonstrated the efficacy of CleaR across diverse configurations of noisy labels. Moreover, our in-depth analysis has demonstrated that CleaR effectively mitigates the underfitting on clean samples of PEFT methods.

%Lastly, we have suggested that CleaR can be beneficially integrated into the existing NLL approaches.
\section{Limitation}
While we have shown that CleaR successfully improves the effectiveness of PEFT on various NLL scenarios, there exists a few limitations. 

\paragraph{Exploration on Different Architectures} Our efforts have been focused on improving the efficacy of PEFT for encoder models, aligning with previous studies \cite{stgn}. Therefore, the applicability of CleaR methods to different architectures (e.g., decoder, encoder-decoder models) remains under-explored in this work. Nevertheless, based on recent evidence suggesting that routing-based PEFT methods can be effectively generalized to various architectures beyond encoder models \cite{adamix, smop}, we believe that CleaR is expected to work well within other architectures. We leave the exploration of this direction as promising future work. 

\paragraph{Computational Overheads} 
The adaptive routing mechanism in CleaR could potentially introduce computational overhead in two main areas: (i) determining the probability of each sample, and (ii) executing the PEFT routing. However, the overhead for (i) can be mitigated by caching the samples' losses during the training phase, eliminating the need for separate procedures to compute the training loss. As for (ii), in comparison to other routing-based methods cited as \cite{smop} that employ parameterized routers, the additional computational costs in CleaR are negligible, as the router in CleaR is non-parametric, and decisions are made through sampling, which streamlines the process.

\section*{Acknowledgment} 
This work was supported by the Basic Research Program through the National Research Foundation of Korea (NRF) grant funded by the Korea government (MSIT) (2021R1A2C3010430) and Institute of Information \& Communications Technology Planning \& Evaluation (IITP) grant funded by the Korea government (MSIT) (No.RS-2019-II190079, Artificial Intelligence Graduate School Program (Korea University)).

% \newpage
\appendix

\begin{center}
\LARGE
\textbf{Appendix}    
\end{center}
\label{sec:appendix}

\begin{figure*}[t]
\centering
\includegraphics[width=\linewidth]{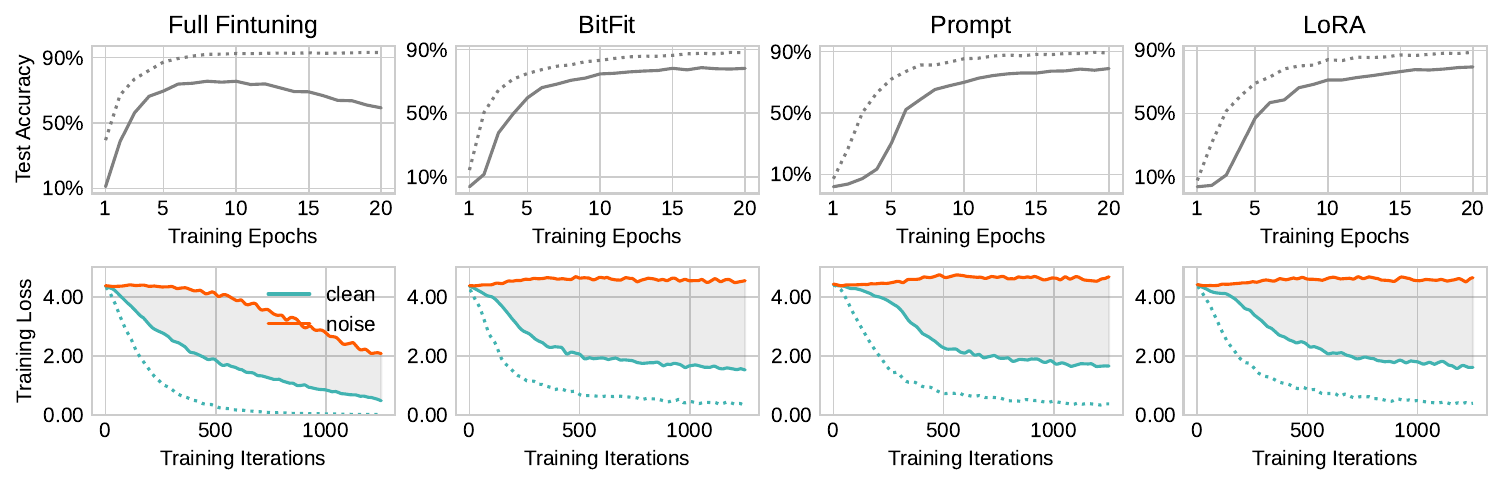}
% \hfill
\caption{Comparison between PEFT methods and full fine-tuning on BANKING77 with symmetric noise (60\%). Dashed lines represent the training accuracy and loss of clean samples on uncorrupted datasets (i.e. only clean samples).}
\label{fig:peft_analysis_bank}
\end{figure*}

\begin{figure*}[t]
\centering
\includegraphics[width=\linewidth]{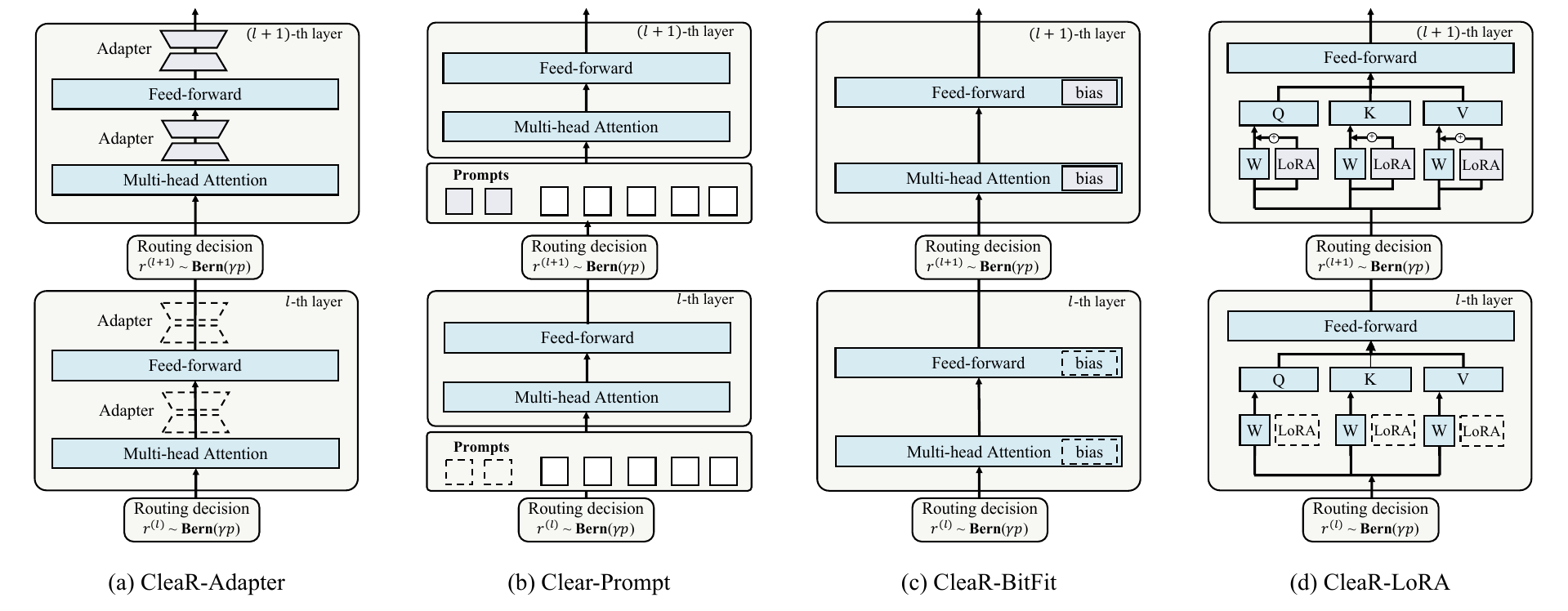}
% \hfill
\caption{Detailed illustration of the CleaR adaptation to PEFT methods (e.g., Adapter, Prompt Tuning, BitFit, LoRA). Dashed lines indicate the unused modules, except for the CleaR$_{\text{BitFit}}$ that uses fixed pre-trained biases.}
\label{fig:clear_model_detail}
\end{figure*}

\section{PEFT Analysis on Different Dataset}
We further conducted the same analysis on a different dataset to validate the generality of our observations. Specifically, we examined PEFT methods on the BANKING77 \cite{banking77} dataset, which focuses on an intent detection task. The results are presented in Figure \ref{fig:peft_analysis_bank}. Similar to our analysis of the SST-5 dataset, we observed a similar trend: (i) PEFT methods exhibit greater robustness than full fine-tuning, and (ii) its limited memorization on noisy label attributes to robustness, although they also inhibit memorization even on clean samples. These findings further support our observations and analysis regarding the robustness of PEFT methods to noisy labels.

\section{Experiments on Additional Datasets}
To further demonstrate the broad applicability of our proposed method, we have evaluated the proposed methods on the TREC and 20Newsgroups datasets. Table \ref{tab:additional_datasets} shows the evaluation results on the two datasets with 60\% symmetric noises. We can observe the similar performance trend with the existing benchmarks, further underscoring the general applicability of the proposed method.

\begin{table}[h]
\centering
\caption{Peak and Average accuracy (\%) on TREC and 20NewsGroups with 60\% symmetric noisy labels. Best results are highlighted in boldface.}

\begin{tabular}{@{}lcccc@{}}
\toprule
\multirow{2}{*}{Method} & \multicolumn{2}{c}{TREC} & \multicolumn{2}{c}{20NewsGroups} \\ \cmidrule(l){2-5} 
                        & Peak.       & Avg.       & Peak.           & Avg.           \\ \midrule
Full Fine-Tuning        & 90.2        & 55.9       & 62.3            & 37.6           \\ \midrule

\multicolumn{5}{l}{\textit{PEFT methods}}                \\ 

Adapter                 & 91.4        & 70.6       & 61.8            & 49.7           \\
BitFit                  & 92.1        & 91.9       & 61.8            & 61.2           \\
LoRA                    & 89.4        & 77.2       & 61.2            & 51.5           \\ \midrule
\multicolumn{5}{l}{\textit{PEFT methods with CleaR (ours)}}                \\ 
CleaR$_{\text{Adapter}}$                 & \textbf{93.9}        & \textbf{92.7}       & 62.9            & \textbf{61.6}           \\
CleaR$_{\text{BitFit}}$                  & 93.3        & 92.6       & 63.0            & 58.2           \\
CleaR$_{\text{LoRA}}$                    & 92.0        & 91.3       & \textbf{63.2}            & 57.4           \\ \bottomrule
\end{tabular}
\label{tab:additional_datasets}
\end{table}

\section{CleaR on Larger Model}
\label{sec:B}
%To further investigate whether CleaR works well in larger models, 
To evaluate how CleaR performs as the model evolves, we compare our CleaR with baselines on BERT-Large, which is 3$\times$ larger than the base-sized model. 
Table \ref{table6} represents the evaluation results on SST-5 with different levels of symmetric noise. We observe CleaR still outperforms baselines on both peak and average accuracy by a large margin. These results demonstrate that our CleaR allows us to improve the model performance on noise label settings regardless of the model sizes. 
%The results are desirable because our CleaR allows us to improve the model performance on noise label settings regardless of model size. 
%특히, 모델의 크기가 커질수록 기존 baseline은 더 낮은 성능을 보이는 것을 확인하였으나, 우리의 CleaR는 여전히 높은 성능을 유지 가능한것을 보였다. 
%CleaR on more realistic and challenging settings, we additionally evaluate CleaR on the scenarios in which noisy labels arise from input features. Table \ref{tab:instance} presents the evaluation results on SST-5 with the instance-dependent noise. Similar to the different noises, CleaR consistently yields the best performance on both peak and average accuracy across different noise ratios, highlighting the validity of CleaR in addressing feature-dependent noise.

\begin{table}[h]
\centering
\caption{Peak and Average accuracy (\%) on SST-5 under the BERT-large.}
\label{table6}
\resizebox{\columnwidth}{!}{
\begin{tabular}{@{}llllll@{}}
\toprule
\multicolumn{1}{c}{\multirow{2}{*}{Method}} & \multicolumn{1}{c}{\multirow{2}{*}{Clean}} & \multicolumn{2}{c}{40\%} & \multicolumn{2}{c}{60\%} \\ \cmidrule(l){3-6} 
\multicolumn{1}{c}{} & \multicolumn{1}{c}{} & \multicolumn{1}{c}{Peak.} & \multicolumn{1}{c}{Avg.} & \multicolumn{1}{c}{Peak.} & \multicolumn{1}{c}{Avg.} \\ \midrule
Full Fine-tuning & 55.1 & 51.3 & 43.7 & 48.6 & 37.0 \\ \midrule

\multicolumn{6}{l}{\textit{PEFT methods}}                \\ 

Adapter & \textbf{55.5} & 50.9 & 46.2 & 49.5 & 40.8 \\
BitFit & \underline{55.4} & 51.7 & 49.8 & 50.5 & 48.2 \\
Prompt & 53.6 & 50.0 & 46.1 & 49.7 & 45.1 \\
LoRA & 54.5 & \underline{52.4} & 48.8 & 49.8 & 46.1 \\ \midrule

\multicolumn{6}{l}{\textit{PEFT methods with CleaR (ours)}}                \\ 

CleaR$_{\text{Adapter}}$ & 55.2 & \textbf{53.1} & \textbf{52.2} & \textbf{53.5} & \underline{49.2} \\
CleaR$_{\text{BitFit}}$ & 54.7 & 51.9 & 50.6 & 51.5 & \underline{49.2} \\
CleaR$_{\text{Prompt}}$ & 53.7 & 52.0 & 50.5 & \underline{52.4} & \textbf{50.5} \\
CleaR$_{\text{LoRA}}$ & 54.5 & 52.2 & \underline{51.2} & 50.6 & 48.8 \\ \bottomrule
\end{tabular}
}
\end{table}

\begin{table}[h]
\centering
\caption{The ratio of training parameters (\%) for each PEFT method. Note that CleaR methods have the same number of trainable parameters.}

\begin{tabular}{@{}cc@{}}
\toprule
Methods & Trainable Parameters (\%) \\ \midrule
Adapter \citeyearpar{adapter} & 0.455\%                   \\
BitFit \citeyearpar{bitfit}  & 0.078\%                   \\
Prompt \citeyearpar{ptuning_v2}  & 0.024\%                   \\
LoRA \citeyearpar{lora}    & 0.111\%                   \\ \bottomrule
\end{tabular}
\label{tab:params}
\end{table}

% \usepackage{booktabs}

\section{CleaR on PEFT methods}
\label{sec:C}
In this paper, we apply CleaR to the existing PEFT methods. Specifically, we select the widely-used and different types of PEFT methods, which include Adapter \cite{adapter}, Prompt Tuning \cite{prefix_tuning}, BitFit \cite{bitfit}, and LoRA \cite{lora}. For each PEFT method, we follow the commonly used setup, and the ratio of trainable parameters is listed in Table \ref{tab:params}. We name the adopted version of each method as CleaR$_{\text{Adapter}}$, CleaR$_{\text{Prompt}}$, CleaR$_{\text{BitFit}}$, and CleaR$_{\text{LoRA}}$. For each method, we provide the graphical procedure in Figure \ref{fig:clear_model_detail}, and its details are as follows:

\paragraph{CleaR$_{\text{Adapter}}$} Adapter \cite{adapter} employs a bottleneck architecture, comprising down- and up-projection matrices with non-linearity applied to the bottleneck feature. These adapters are positioned above the multi-head self-attention and position-wise feed-forward layers. In the CleaR method, we group these two adapters and determine their usage through a sampling process.

\paragraph{CleaR$_{\text{Prompt}}$} We adopt P-Tuning v2 \cite{ptuning_v2} as a representative method for prompt tuning. This approach simply appends trainable prompts to the original input embeddings and fine-tunes these appended prompt embeddings during the fine-tuning phase. In the CleaR method, we group these prompt embeddings and decide whether or not the prompts are appended to the input. As a result, in CleaR$_{\text{Prompt}}$, the input length can vary based on the routing decision.

\paragraph{CleaR$_{\text{BitFit}}$} Different to other PEFT methods that introduce trainable weights (e.g., adapters, prompts), BitFit \cite{bitfit} fine-tunes the existing biases in the PLMs. To apply routing to BitFit, we store the pre-trained biases, and the routing mechanism determines whether the training model utilizes the pre-trained (fixed) biases or the trainable biases.

\paragraph{CleaR$_{\text{LoRA}}$} LoRA \cite{lora} employs a bottleneck architecture similar to the adapter. However, the bottleneck in LoRA is situated within the weights for the attention matrices. In the CleaR method, we decide whether the linear weights for the attention mechanism (e.g., query, key, and value weights) incorporate trainable weights or not.

\section{Detailed Process for Generating Noisy Labels}\label{appendix:noise_process}
We detail the process for generating noisy labels when the noise rate $p$ is given. 
\begin{itemize}
    \item \textbf{Symmetric noise:} To generate this noise, we create the noise transition matrix $T \in \mathbbm{R}^{C \times C}$ where $C$ is the number of classes. We then set a value of $p$ to its diagonal elements and distribute the remaining probability $(1-p)$ to other non-diagonal elements. Based on the probability in the matrix, we flip the labels in training samples.
   
    \item \textbf{Asymmetric noise:} Similar to the symmetric noise, we create the noise transition matrix $T \in \mathbbm{R}^{C \times C}$ where $C$ is the number of classes. We then set a value of $p$ to its diagonal elements and assign the remaining probability $(1-p)$ to the next element of the diagonal values to implement single-flip noise \cite{selfmix}. Based on the probability in the matrix, we flip the labels in training samples.

    \item \textbf{Instance-dependent noise:} To generate instance-dependent noise, we first pre-train the classifier on the original dataset. We then select the two classes, which are the most confident $u$ and the second most confident classes $s$, and calculate the distance of decision boundaries between the classes, i.e., $[f_{u}(x)-f_{s}(x)]^2$. Based on the distance, we define the noise function as $\tau = -\frac{1}{2}[f_{u}(x)-f_{s}(x)]^{2} + \frac{1}{2}$. A smaller distance between the classes results in a larger flipping probability to the second most confident class $s$. Lastly, to control the degree of noisy labels, we multiply $\tau$ by a certain constant factor such that the final proportion of noise matches the pre-defined noise probability.
\end{itemize}

\section{Implementation Details and Setups}\label{appendix:implement}
In this section, we detail to implement the baselines and our CleaR on various tasks.
\paragraph{PEFT implementation. } We compare our CleaR with four strong baselines, which include Adapter \cite{adapter}, LoRA \cite{lora}, BitFit \cite{bitfit}, and Prompt-tuning \cite{ptuning_v2}. Specifically, we set the bottleneck dimension $r$ for the Adapter and LoRA as 16 and 4, respectively. For LoRA, we only apply LoRA weights on query and value attention weights. Moreover, we fine-tune all bias parameters in transformer blocks for BitFit. For Prompt-tuning, we set the fixed length as 20 for prompts in each transformer layer, following P-tuning v2 \cite{ptuning_v2}.
\paragraph{Hyper-parameters. } 
For the hyper-parameters to fine-tune in CleaR, we select the best warmup epoch in $[3,10]$ and clean probability weights $\gamma$ in $[0.5, 1]$ corresponding to each PEFT method and task. We also use the number of forward $N = 5$ for constructing ensemble predictions on consistency regularization. we use the consistency regularization coefficient $\lambda = 1$ for all experiments.
For other settings, we use Adam optimizer \cite{adam} with $\beta_1 = 0.9$ and $\beta_2 = 0.999$. We also train all models using a batch size of 32 and sweep the learning rates in \{1e-4, 2e-4, 3e-4, 4e-4, 5e-4\} for PEFT methods. For full fine-tuning, we select the best learning rates in \{1e-5, 2e-5, 3e-5, 4e-5, 5e-5\}. All models are fine-tuned for 20 epochs.% The exact values are in Table 6.
%Other settings like learning rate, optimizer, and batch size keep the same for all the methods and tasks.

\paragraph{Hardware Details. } We train all our models using four RTX 3090 GPUs. We utilize mixed-precision training \cite{mixed_precison} to expedite the training procedure. All the implementations are performed using the PyTorch framework.

\section{Details for other NLL Methods}\label{appendix:nll}

We compare our CleaR with the following NLL methods: 
\paragraph{Co-teaching. } Co-teaching \cite{co-teaching} trains two models simultaneously and lets each model select clean training samples (i.e., small-loss instances) for training each other model. Co-teaching framework gradually drops the noisy samples to prevent overfitting. To this end, they require an estimation of the noise level ($\tau$), warm-up steps ($T_k$), and coefficient ($c$). We set the noise level $\tau$ as the ground truth noise ratio. We vary the warm-up steps $T_k$ in \{1500, 2000, 2500, 3000, 3500\}, and select the best coefficient $c$ from search range of $[1, 2]$ for each fine-tuning method.
\paragraph{SELC. } SELC \cite{selc} trains the model using ensemble prediction based on historical model outputs to correct the noisy labels. Specifically, they first train their models with given labels until the turning point $T$, which represents the model would start overfitting on noisy levels. And then they combine the given labels with ensemble predictions with momentum $\alpha$ as the target. We estimate the turning point $T$ by leveraging metrics, following \cite{selc}, corresponding to each fine-tuning method. We also select the best momentum $\alpha$ by searching parameters from $[0.5, 1]$.
\paragraph{STGN. } STGN \cite{stgn} trains the model by reducing the disturbance on correct samples and increasing the perturbation on corrupted ones. Specifically, they utilize the standard deviation $\sigma_f$ to perturb the gradient of loss and forgetting events threshold $\lambda_f$ to separate corrupted data from correct ones. For full fine-tuning, we use the same setup in \cite{stgn}. For adapter-based tuning, we set $\sigma_{max} = 2 \sigma_f$ and select the $\sigma_f = 7e-4$ and $\lambda_f = 4$. 

We implement other NLL methods based on standard BERT with reference to their public code and make comparisons under the same setting. For other hyper-parameters (i.e., learning rate, and warmup epochs), we select the optimal values by searching on the same parameter space with baselines and CleaR. %The detailed hyper-parameters are shown in Table 7. 

\end{document}